# Phonological Derivation in Optimality Theory*

T. Mark Ellison

Centre for Cognitive Science, University of Edinburgh
2 Buccleuch Place, Edinburgh EH8 9LW, Scotland

**Summary:** Optimality Theory is a constraint-based theory of phonology which allows constraints to conflict and to be violated. Consequently, implementing the theory presents problems for declarative constraint-based processing frameworks. On the basis of two regularity assumptions, that sets are regular and that constraints can be modelled by transducers, this paper presents and proves correct algorithms for computing the action of constraints, and hence deriving surface forms.

## INTRODUCTION

Recent years have seen two major trends in phonology: theories have become more oriented around constraints than transformations, while implementations have come to rely increasingly on finite state automata and transducers. This paper seeks to build a bridge between these trends, showing how one constraint-based theory of phonology, namely Optimality Theory, might be implemented using finite-state methods.

The paper falls into three main sections. The first describes Optimality Theory and its restriction to constraints which can only make binary distinctions in harmony. The second part covers the formalisation of the evaluation of harmony, including the simplifying assumptions that the set of candidate forms must initially be regular, and that the action of each constraint in assigning harmony also be regular. The third section presents algorithms for (i) defining the product of automata modelling constraints, (ii) finding the optimal level of harmony of a set of candidates and (iii) culling suboptimal candidates. The last two algorithms are proved correct, and some worst-case complexity results are given. The paper concludes with a discussion of the work.

## OPTIMALITY THEORY

Optimality Theory (OT) is a constraint-based theory of phonology, developed by Prince and Smolensky (1993) (hereafter, this work will be referred to as P&S) and is now being used by a growing number of phonologists (Ito and Mester 1993, McCarthy and Prince 1993, McCarthy 1993). It differs from declarative phonology (Bird 1994, Scobbie 1991, Bird and Ellison 1994) in that its constraints are violable and can conflict, with the conflicts resolved by an ordered system of defaults[1]. Declarative phonology evaluates candidate forms[2] on a binary scale: whether they are accepted by a constraint system or not. In contrast, OT assigns a ranking to all of the candidate realisations of a word, calling the scale a measure of *harmony*. All of the candidates which show the maximal amount of harmony are accepted by the constraint system, and others are rejected. A derivation in OT consists of an original candidate set produced by a function called GEN, and the subsequent application of constraints to reduce the candidate set, eliminating non-optimal candidates and preserving those with the greatest harmony. At no stage can a constraint eliminate all candidates.

Each constraint assigns to each candidate a list of *marks*. These marks may, for instance, tag segments as regular or exceptional. The marks are values on the harmony scale, and are totally ordered: for any two marks $a$ and $b$, either $a$ is more harmonic than $b$ (symbolically, $a \succ b$) or the reverse. In the list assigned to a candidate, however, the same mark may occur many times. To compare the harmony of two candidates with regard to a given constraint, their respective lists of marks are sorted into increasing order of harmony[3]. The lists are then compared first-to-last componentwise. The more harmonic candidate has the more harmonic value at the first point where the lists differ. The empty list always has the same harmony as the most harmonic mark on the harmony scale, common to all constraints, which we will call the *zero* mark, and write as $\emptyset$[4]. Constraints which only use two different marks are called *binary* constraints. For binary constraints, the evaluation of harmony is a simple affair. The candidate with the fewest non-zero marks is preferred.

Consider, for example, the binary constraint ONS(P&S:25). This constraint discourages nuclei without onsets when selecting between different syllabifications. Two syllabifications of the Arabic segmental sequence **alqalamu** are shown in (1), with syllables demarcated by parentheses. The nuclei are always the vowels. The dishar-

---

*This research was funded by the U.K. Science and Engineering Research Council, under grant GR/G-22084 *Computational Phonology: A Constraint-Based Approach*. I am grateful to Steven Bird, Ewan Klein and Jim Scobbie for their comments on an earlier version of this paper.

[1] Ellison (1994) offers a formal analysis of the use of defaults in Optimality Theory.

[2] In constraint-based theories, constraints impose limits on possible realisations of objects, such as words or sentences. A candidate is a tentative realisation which is yet to be tested against the constraints.

[3] Early in their technical report, P&S introduce one constraint, HNUC, which requires sorting into the reverse order. Later in the same work they replace this constraint with a number of binary constraints with the usual ordering.

[4] The zero mark is not part of P&S's account. They are not explicit about comparison of lists of marks of unequal length except in the binary case. In that case, their definitions have the same consequences as those described here.

monic mark *L* indicates on onsetless nucleus, the harmonic (zero) mark ∅ is used for other segments.

(1)
| syllabification | marks | sorted |
|---|---|---|
| (al)(qa)(la)(mu) | L∅∅∅∅∅∅∅ | L∅∅∅∅∅∅∅ |
| (alq)(al)(am)(u) | L∅∅L∅L∅L | LLLL∅∅∅∅ |

In this example, the sorted lists of marks differ in the second position with the first candidate, **(al)(qa)(la)(mu)**, being the more harmonic of the two.

When there is more than one constraint, we must consider not only the ordering of marks assigned by one constraint, but the ordering of marks from different constraints. In OT, constraints are placed in a total order ($C_1 \succ\succ C_2$), and all non-zero marks of higher-ranked constraints ($C_1$) are less harmonic than all non-zero marks of lower-ranked constraints ($C_2$). In effect, this means that higher-ranked constraints have priority in eliminating candidates. For all constraints, however, the zero mark has the same, maximally harmonic, value.

## binarity

So far we have considered a general class of constraints including non-binary constraints. As it happens, non-binary constraints can often be replaced by binary constraints. Binary constraints are those which only assign two marks: the zero mark, and one other.

In the simplest case, restating a constraint in a logically equivalent form can transform a non-binary constraint into a binary constraint. The constraint family EDGEMOST is defined by P&S(p35) as (2).

(2)     EDGEMOST($\phi$; E; D).
        The item $\phi$ is situated at the edge E of domain D.

This definition covers a family of constraints depending on the instantiations of the arguments: E is either left (L) or right (R), domain may be syllable, foot or word, and $\phi$ can be any phonological object, such as stress or an affix.

According to P&S, constraints of this form are nonbinary, returning as their marks the distance of their objects from the designated edge of domain. The greater the distance, the less harmonic the mark. Constraints of this kind can, however, be replaced by logically equivalent binary constraints (3).

(3)     NOINTERVENING($\phi$; E; D).
        There is no material intervening between $\phi$ and edge E of domain D.

This form of constraint assigns a disharmony mark to each item intervening between $\phi$ and edge E. The more material lying between $\phi$ and E, the greater the number of marks and so the lower the harmony value.

Other types of non-binary constraints can be converted into hierarchies (ordered sequences) of binary constraints. Suppose a constraint C produces N different kinds of marks. Applied to a candidate form $c$, this constraint produces a list C($c$) of marks. Now define a function $f$ which takes a list of marks, $l$, and a mark type $m$, and replaces all marks in $l$ which are different from $m$ by the zero mark ∅, and then re-sorts the list. So with the marks $2 \prec 1 \prec \emptyset$, then $f(221\emptyset, 2)$ is $22\emptyset\emptyset$ and $f(221\emptyset, 1)$ is $1\emptyset\emptyset\emptyset$. If the marks generated by C are $\emptyset = m_1 \succ m_2 \succ .. \succ m_N$, then C can be replaced by constraints $C_{i,i=1..N-1}$ such that $C_i(c) = f(C(c), i)$ subject to the ordering $C_i \succ\succ C_j$ if $i > j$.

To see the equivalence of the single non-binary constraint with the family of binary constraints, let us look at the comparison of some candidate forms. Using the three-valued constraint of the earlier example, suppose candidates M, N and P are assigned mark lists $1\emptyset2$, $21\emptyset12$ and $\emptyset122$ respectively. Sorted, these lists become $21\emptyset$, $2211\emptyset$ and $221\emptyset$. Comparing these lists, we arrive at the harmony ordering M $\succ$ P $\succ$ N.

Now, let us apply the corresponding binary constraints. The first and dominant constraint preserves only 2s in the mark list, the second preserves only the mark 1. The two lists of marks for M, N and P are $2\emptyset\emptyset$ and $1\emptyset\emptyset$, $22\emptyset\emptyset\emptyset$ and $11\emptyset\emptyset\emptyset$, and $22\emptyset\emptyset$ and $1\emptyset\emptyset\emptyset$, respectively. By the ordering of the constraints, we know that $2 \prec 1$ still, and so merging the two lists of marks for each candidate gives $21\emptyset\emptyset\emptyset\emptyset$, $2211\emptyset\emptyset\emptyset\emptyset\emptyset$ and $221\emptyset\emptyset\emptyset\emptyset\emptyset$. Apart from the trailing ∅s, these are identical to the marks assigned by the single constraint, and so lead to the same ordering: M $\succ$ P $\succ$ N.

So all constraints which use a finite alphabet of marks, and some which do not, such as EDGEMOST constraints, can be translated into binary constraints or a finite sequence of binary constraints. Consequently, formalising binary constraints and their interaction will be enough to capture the bulk of constraints in OT.

## FORMALISATION

The formalisation of OT developed here makes uses three idealising assumptions (4).

(4)     1. All constraints are binary.
        2. The output of GEN is a regular set.
        3. All constraints are regular.

We have already seen that most non-binary constraints can be recast as binary constraints or families of binary constraints. Unfortunately, P&S are not explicit about whether there are other unbounded non-binary constraints (like EDGEMOST) — there may be some which cannot be recast as binary constraints. Assumption 1 is, therefore, an idealisation imposing a slight limitation on the theory.

### regular gen

The second assumption requires that the output of GEN be regular. Recall that GEN is the function which produces the initial set of candidate forms which is reduced by the constraints. In other words, the set of candidates must be initialised to a set which can be defined by a regular expression, or, equivalently, by a finite-state automaton (FSA).

As an example, (5) shows a regular expression giving a subset of the candidate syllabifications of **alqalamu** according to the syllabification rules of P&S(p25). The set does not include all candidates; for clarity I have omitted partial syllabifications in which segments have not been assigned a syllabic role, and completely empty syllables. The set does include syllabic slots which do not correspond to segments. In such slot-segment pairs, the empty segment is written as 0.

(5)
$$\left\{\begin{matrix}O\\0\end{matrix}\Big|\right\} NCON \left\{\begin{matrix}a & l & q & a\end{matrix}\right. \left\{\left\{\begin{matrix}C\\0\end{matrix}\Big|\right\}\begin{matrix}O\\l\end{matrix}\Big|\begin{matrix}C\\l\end{matrix}\Big|\left\{\begin{matrix}O\\0\end{matrix}\Big|\right\}\right\}$$
$$\begin{matrix}N\\a\end{matrix} \left\{\left\{\begin{matrix}C\\0\end{matrix}\Big|\right\}\begin{matrix}O\\m\end{matrix}\Big|\begin{matrix}C\\m\end{matrix}\Big|\left\{\begin{matrix}O\\0\end{matrix}\Big|\right\}\right\} \begin{matrix}N\\u\end{matrix} \left\{\begin{matrix}C\\0\end{matrix}\right\}$$

The brackets cover disjunctions of terms separated by vertical bar |, while concatenation is expressed by juxtaposition. The vertical pairs of symbols are the complex labels used on arcs in the corresponding automaton. The three syllabic slot types are onset (O), nucleus (N) and coda (C). As a regular expression, (5) captures 64 different possible syllabifications of the sequence **alqalamu**. For example, the syllabification **(al)(qal)(am)(u)** is accepted by the (5), while **(alq)(al)(am)(u)** is not.

### regular constraints

The third assumption imposes regularity on constraints. A constraint is regular if there is a finite-state transducer[5] (FST) which assigns the same list of marks to a candidate form that the constraint does. Since we are only dealing with binary constraints, the transducer will associate with each component of the candidate one of the two harmonic values $\epsilon \prec \emptyset$. Such transducers can be expressed as regular expressions over pairs of phonological material and marks.

P&S (p25) use two constraints, FILL (6) and ONS (7), to account for the limits on epenthesis in Arabic. Epenthetic material arises when syllabic slots which are not occupied by segments are realised. Here the marks are given on the right hand side of the colon in each pair. Here $\epsilon$ is the disharmonic mark, and $\emptyset$ the more harmonic zero mark.

(6) FILL. Syllable positions are filled with segmental material.

(7) ONS. Every syllable has an onset.

These two constraints can be readily translated into regular expressions, using the abbreviatory notations: $\overline{N}$ for onset or coda, $\overline{0}$ for segmental material and • for anything. The transducers for FILL and ONS are defined by the regular expressions in (8) and (9) respectively.

(8)
$$\left\{\begin{matrix}\epsilon & \emptyset \\ \bullet & \bullet \\ 0 & \overline{0}\end{matrix}\right\}^{\star}$$

This transducer marks with $\epsilon$ every syllabic slot associated with an empty (0) segment.

(9)
$$\left\{\begin{matrix}\emptyset & \emptyset & \emptyset & \epsilon \\ OO & \overline{N} & N \\ \bullet \bullet & \bullet & \bullet\end{matrix}\right\}^{\star}$$

---

[5]A finite state transducer is an FSA which is labelled with pairs of values. In this case, the pairs will combine phonological information with constraint marks.

This transducer is non-deterministic, producing more than one sequence of marks for a given input. All nuclei preceded by an onset are marked with $\emptyset$ and with $\epsilon$. All other segments segments are marked as $\emptyset$. The multiple evaluations of candidates is not a problem: candidates will survive so long as their best evaluation is as good as the best of any other candidate.

### linearity

The reader may be concerned that the regularity constraint imposes undue restrictions of linearity on the candidate forms, and, in doing so, vitiates the phonological advantages of non-linear representations. This is not the case. Bird and Ellison (1992,1994) have shown that it is possible to capture the semantics of autosegmental rules and representations using FSAs. The output of GEN, therefore, may correspond to a set of partially specified autosegmental representations, and still be interpreted as a regular set.

### candidate comparison

For single binary constraints, the harmony of candidates is compared as sorted lists over the alphabet containing $\epsilon$ and $\emptyset$, where $\emptyset$ has the higher harmony, the same, in fact, as the empty list. Consequently, the results of comparing lists of these marks is identical with comparing $\#\epsilon.h(\epsilon)$ where $\#\epsilon$ is the number of times $\epsilon$ occurs in the list, and $h(\epsilon)$ is the constant quantity of harmony assigned to $\epsilon$. As $\emptyset$ has the same harmony as the empty list, $h(\emptyset)$ must be zero. As $\epsilon \prec \emptyset$, comparison is preserved if $h(\epsilon) < 0$, so we set $h(\epsilon) = -1$. If the arcs in the transducer are labelled with $-1$ and $0$ instead of $\epsilon$ and $\emptyset$, then the harmony of a candidate can be evaluated by just adding the numbers along the corresponding path in the constraint transducer. The greater the (always non-positive) result, the more harmonic the candidate.

Just as we can measure harmony relative to a single constraint with a single integer, we can measure the harmony relative to an ordered hierarchy of constraints with an ordered list of integers. The list of integers corresponds one-to-one to the constraints in decreasing order of dominance. Each integer maintains information about the number of $\epsilon$ values of the corresponding constraint in the evaluation of the candidate. A candidate with the list $(-2, -1)$ violates the first constraint twice and the second once: the corresponding sorted list of harmony marks is $221$.

Lists of this form can be compared just like lists of harmony marks. The first integer is the most significant and the last the least. The greater of two lists is the one with the higher value at the most significant point of difference. For example $(-10, -31, -50)$ is more harmonic than ($>$) $(-10, -34, -12)$. Lists of integers can be accumulated like single integers using componentwise addition.

We can generalise transducers from denoting single constraints to denoting hierarchies of constraints: from translating candidates into sequences of $\{\emptyset, \epsilon\}$ or $\{0, -1\}$ marks to transducers from candidates to sequences of lists of integers, each integer drawn from $\{0, -1\}$. Summing the lists along a path gives a harmonic evaluation of the corresponding candidate.

Let us call the length of the integer list the *degree* of the transducer. The output of GEN is an automaton — a transducer without marks — and so corresponds to a transducer of degree 0. The transducer for a single constraint needs only a single binary distinction for its marks, so a degree 1 transducer suffices. In general, the number of binary constraints that a transducer encodes will equal its degree. The next section looks at how transducers of single constraints or small hierarchies can be combined into single transducers for larger hierarchies.

## ALGORITHMS

### product

We have seen how a single constraint can be regarded as a transducer from candidate segments into a singleton list of integers, and further that multiple constraints can be evaluated using longer lists of integers. Combining these two notions into an extended version of the automaton product operation allows us to build up transducers capturing a hierarchy of constraints from single constraint transducers.

The product operation is easier to describe when transducers are thought of in terms of automata rather than regular expressions. For brevity, then, the algorithms will be phrased in terms of the states and arcs of an automaton, while, for clarity, regular expressions will be used to present the inputs and outputs of examples.

The pseudocode for the standard automaton product operation appears in (10). As the initial states of any automaton can be identified with each other without affecting the language recognised, and similarly the final states, we will assume that there is only a single initial state (I) and final state (F) in each automaton. In this pseudocode, semicolons are followed by comments.

(10)   Product($\mathcal{A},\mathcal{B}$):
    1   make (I$_\mathcal{A}$,I$_\mathcal{B}$) initial in $\mathcal{A} \times \mathcal{B}$
    2   make (F$_\mathcal{A}$,F$_\mathcal{B}$) final in $\mathcal{A} \times \mathcal{B}$
    3   **for each** arc from x to y in $\mathcal{A}$ labelled $\mathcal{M}$
    4     **for each** arc from z to t in $\mathcal{B}$ labelled $\mathcal{N}$
    5       **if** $\mathcal{M} \cap \mathcal{N} \neq \emptyset$
    6       **then** add arc from (x,z) to (y,t) to $\mathcal{A} \times \mathcal{B}$
    7         labelled $\mathcal{M} \cap \mathcal{N}$

The pseudocode in (10) applies to two automata $\mathcal{A}$ and $\mathcal{B}$, over the same alphabet, and constructs their product $\mathcal{A} \times \mathcal{B}$, an automaton which accepts only those strings accepted by both $\mathcal{A}$ and $\mathcal{B}$. Each combination of arcs, one from $\mathcal{A}$ and one from $\mathcal{B}$, which could be traversed while reading the same input, that is, an input in the intersection $\mathcal{M} \cap \mathcal{N}$ of the labels of the two arcs, defines an arc in the product automaton.

To make the product mimic the combination of constraints in OT, we need to introduce an asymmetric operation on the lists of marks: concatenation. Each arc in each automata passed to this product operation is labelled not only with a set of possible phonological segments, but also a list of harmony marks. When two arcs are combined, these lists are concatenated. The pseudocode for this augmented product operation appears in (11).

(11)   AugmentedProduct($\mathcal{A},\mathcal{B}$):
    1   make (I$_\mathcal{A}$,I$_\mathcal{B}$) initial in $\mathcal{A} \times \mathcal{B}$
    2   make (F$_\mathcal{A}$,F$_\mathcal{B}$) final in $\mathcal{A} \times \mathcal{B}$
    3   **for each** arc from x to y in $\mathcal{A}$ labelled $\mathcal{M}:\mu$
    4     **for each** arc from z to t in $\mathcal{B}$ labelled $\mathcal{N}:\nu$
    5       **if** $\mathcal{M} \cap \mathcal{N} \neq \emptyset$
    6       **then** add arc from (x,z) to (y,t) to $\mathcal{A} \times \mathcal{B}$
    7         labelled $\mathcal{M} \cap \mathcal{N}:\mu.\nu$
      $\mu.\nu$ is the concatenation of $\mu$ and $\nu$.

Because concatenation is not a symmetric operation, the augmented product does not commute: $\mathcal{A} \times \mathcal{B}$ does not assign the same marks to candidate forms as $\mathcal{B} \times \mathcal{A}$. The difference in interpretation is that $\mathcal{A} \times \mathcal{B}$ regards all constraints in $\mathcal{A}$ as higher priority than all constraints in $\mathcal{B}$, whereas $\mathcal{B} \times \mathcal{A}$ instantiates the reverse ordering.

The augmented product operation provides a way of combining two constraints into a single transducer. As an example, (12) is the product of the transducers corresponding to the constraints ONS (9) and FILL (8) in that order.

(12)
$$\left\{ \left\{ \begin{matrix} 0\overline{1} & 00 \\ O & O \\ 0 & \overline{0} \end{matrix} \right\} \left\{ \begin{matrix} 0\overline{1} & 00 \\ N & N \\ 0 & \overline{0} \end{matrix} \right\} \mid \left\{ \begin{matrix} 0\overline{1} & 00 \\ \overline{N} & \overline{N} \\ 0 & \overline{0} \end{matrix} \right\} \left\{ \begin{matrix} \overline{11} & \overline{1}0 \\ N & N \\ 0 & \overline{0} \end{matrix} \right\} \right\}^\star$$

The product is the crucial operation for implementing OT. The product of the regular expression or automaton produced by GEN with all of the constraints in order produces a transducer encoding the harmony evaluations of all candidates. Let us call it the *surface* transducer. To evaluate the harmony of any fully specified candidate, we need only follow the corresponding paths in the surface transducer accumulating the integer lists associated with each arc. The path with the greatest total harmony is the crucial one for deciding whether the candidate is optimal or not.

The surface transducer which is the product of the candidate syllabifications of **alqalamu** with the constraints ONS and FILL, in that order, is shown in (13).

(13)
$$\left\{ \begin{matrix} 0\overline{1}\bullet 0 & \overline{1}0 \\ O\ N & N \\ 0\ a & a \end{matrix} \right\} \begin{matrix} 0000\bullet 0 \\ C\ O\ N\ — \\ l\ q\ a \end{matrix}$$

$$\left\{ \left\{ \begin{matrix} 0\overline{1} \\ C \\ 0 \end{matrix} \right\} \begin{matrix} 00\bullet 0 \\ O\ N \\ l\ a \end{matrix} \middle| \begin{matrix} 00 \\ C \\ l \end{matrix} \left\{ \begin{matrix} 0\overline{1}\bullet 0 \\ O\ N \\ 0\ a \end{matrix} \middle| \begin{matrix} \overline{1}0 \\ N \\ a \end{matrix} \right\} \right\} —$$

$$\left\{ \left\{ \begin{matrix} 0\overline{1} \\ C \\ 0 \end{matrix} \right\} \begin{matrix} 00\bullet 0 \\ O\ N \\ m\ u \end{matrix} \middle| \begin{matrix} 00 \\ C \\ m \end{matrix} \left\{ \begin{matrix} 0\overline{1}\bullet 0 \\ O\ N \\ 0\ u \end{matrix} \middle| \begin{matrix} \overline{1}0 \\ N \\ u \end{matrix} \right\} \right\}$$

### harmony of substrings

In OT, only the candidates with maximal harmony survive to the surface; non-optimal candidates are eliminated. To implement this part of the derivation, we need to remove all paths from the surface transducer which do not accumulate optimal values of harmony. The algorithms in this section and the next are designed to achieve this task, and will be proven to do so.

The first algorithm (14) assigns to every state N the harmony value of the optimal path to it from the initial state, storing this value in the field *harmony*(N). Since there is only a single final state, F, harmony(F) will contain the harmony evaluation of all optimal candidates.

(14)   LabelNodes(transducer):
```
   1   for each state n in transducer
   2      harmony(n) undefined
   3   harmony(I) ← 00...0  I is the initial state
   4   list ← { I }
   5   while list is not empty
   6      expand m begins
   7      m ← most harmonic state in list
   8      delete m from list
   9      for each arc a:m→n from m
  10         if harmony(n) < harmony(m) + harmony(a)
  11            delete n from list
  12            harmony(n) ← harmony(m) + harmony(a)
  13            insert n in list
  14         else if harmony(n) undefined
  15            harmony(n) ← harmony(m) + harmony(a)
  16            insert n in list
```

The algorithm sets the harmony of the inital state to zero, and places the initial state in an otherwise empty list. The most optimal member in the list is expanded (lines 6-15) and removed from the list. When a state is expanded, all of the arcs from it are examined in turn. If any of them point to states with undefined harmony values, the harmony of the state being expanded, and of the arc, are used to calculate the harmony value of the other state and it is added to the list. If the arc points to a state with a defined harmony value, the harmony value of the better path is retained by that state, and its position in the sorted list adjusted appropriately.

If the list is kept sorted, inserting each new state in order of the value of its harmony field then, in the worst case, $o(\log |states|)$ comparisons of harmony values will need to be done for an insertion into the list where $states$ is the set of states in the transducer and $arcs$ the set of arcs. As each state is expanded only once, each arc is examined only once. So $|arcs|$ forms an upper bound on the number of insertions that need to be done. The single comparison on line 9 is insignificant in relation to the comparisons used in insertion. So an upper bound on order of the worst case execution of this algorithm is $o(|arcs| \log |states|)$ comparisons.

It is not obvious that this algorithm will, in fact, label each state with the harmony of the optimal path to it, so a proof follows.

(15)   **Lemma.**  When state M is being expanded (lines 6-15), the true harmony value of the optimal path to M, namely $h(M)$, and the computed value, harmony(M), are equal, if the same is true for all previously expanded states.

**Proof.**   **Case $\leq$.** Suppose the lemma is false, and that $h(M)$>harmony(M). Then there is an optimal path $p.a.q$ where $p$ is a (possibly null) path, $a$ is an arc from an already expanded state R to an unexpanded state S and $q$ is another (possibly null) path. There will always be such a path as $M$ is reachable from the initial state, and the initial state is the first one expanded. This path is optimal, so $h(M) = h(R) + h(a) + h(q)$ which in turn is less than or equal to $h(R) + h(a)$ as $h$ is always non-positive. Putting this inequality together with the supposition of the lemma that harmony and $h$ match for all expanded nodes, gives the following inequality:

$$\begin{aligned} harmony(S) &\geq harmony(R) + harmony(a) \\ &= h(R) + h(a) \\ &> harmony(M) \end{aligned}$$

A lower bound for harmony(S) was set when R was expanded. As R is already expanded $h(R)$=harmony(R), and consequently harmony(S)>harmony(M) which contradicts the minimality of the choice of M (line 6 of algorithm (14)). Thus $h(M) \leq$ harmony(M).

**Case $\geq$.** If M is in list, then harmony(M) must be defined and set at a value $\leq h(M)$.

Thus the equality of harmony(M) and $h(M)$, and the lemma. When M is the initial state I, the result follows immediately from line 3 which sets harmony(I) to zero.   □

(16)   **Theorem.**  After the application of LABELNODES, for all states N on which harmony(N) is defined, harmony(N) is the harmony of the optimal path to N.

**Proof.**   By the lemma and induction on the sequence of expansion of states.   □

We can mimic the labelling of nodes in the transducer with harmonic evaluations by labelling disjuncts in regular expressions with harmonic values. The value of a whole disjunction is most harmonic value amongst the disjuncts. As before, the harmonic evaluations are added during concatenation. The evaluations for the surface transducer (13) of **alqalamu** are shown in (17).

(17)   $\left\{ \begin{matrix} 0\overline{1}\bullet 0 & |\overline{1}0 \\ O\ N & |N \\ 0\ a & |a \end{matrix} \right\} \begin{matrix} 0000\bullet 0 \\ C\ O\ O \\ l\ q\ a \end{matrix}$ —
   $\underbrace{\phantom{xxxxxx}}_{0\overline{1}}$
   $\underbrace{\phantom{xx}}_{0\overline{1}}$
   
   $\left\{ \underbrace{\left( \begin{matrix} 0\overline{1} \\ C \\ 0 \end{matrix} \right) \begin{matrix} 00\bullet 0 \\ O\ N \\ l\ a \end{matrix}}_{00} \middle| \underbrace{\begin{matrix} 00 \\ C \\ l \end{matrix} \left( \begin{matrix} 0\overline{1}\bullet 0 & |\overline{1}0 \\ O\ N & |N \\ 0\ a & |a \end{matrix} \right)}_{0\overline{1}} \right\}$ —
   $\underbrace{\phantom{xx}}_{00}$

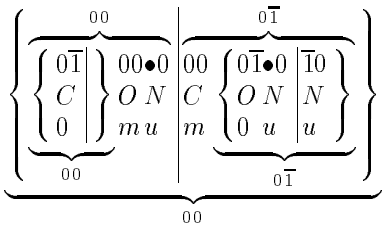

The evaluation of the optimal path in the transducer is (0,-1).

**pruning**

Having determined the harmony value of the optimal path to the final state, and others, it only remains to remove suboptimal paths. As it happens, this can be easily done by removing all arcs which cannot occur in an optimal path (18).

(18)   Prune(transducer):
   1   **for each** arc a:n→m of transducer
   2      **if** harmony(a) + harmony(n) < harmony(m)
   3      **then** delete a

If the sum of the harmony of an arc, and the harmony of the optimal path to the state it comes from, is less than the harmony of the state the arc goes to, then that means that there is a more optimal path to the second state which will always be preferred. Consequently this arc can never be part of an optimal path. It is, therefore, safe and appropriate to delete it.

The complexity, in the number of comparisons performed, of this algorithm is identical to the number of arcs in the transducer. This is of lower order than the worst-case complexity for LABELNODES, so the complexity of the combined algorithm is still $o(|arcs|\log|states|)$ comparisons.

It is not immediately obvious that the only paths which can be formed by the remaining arcs are optimal. This is, however, the case.

(19)   **Theorem.** After the application of LABELNODES and PRUNE there are no non-optimal paths from the start state to any state.

**Proof.** By induction on the length of the path.
$P(n)$ = After the application of LABELNODES and PRUNE there is no non-minimal path of length $n$ from the initial state to any other state.
**Base case.** $P(0)$ is trivially true, as there is only a unique path of length zero.
**Step.** Assume $P(k)$ is true. Suppose we have a non-optimal path of length $k + 1$. By the assumption, this must consist of an optimal path of length $k$ followed by a non-optimal arc $a$ from M to N. $a$ would have been deleted unless harmony(M)+harmony($a$)≥harmony(N). But, by theorem (16), harmony(N) is the harmony of the optimal path to N. So harmony(M)+harmony($a$)≤harmony(N), and the path must be optimal. This contradicts our supposition, and so $P(k + 1)$ is true.
The theorem follows by induction.   □

Consequently, the only paths from the initial state to the final state will be optimal and define optimal candidates.

The regular expression corresponding to the culled automaton describing the syllabifications of **alqalamu** appears in (20). It includes only a single candidate syllabification of the sequence.

(20)   $0\bar{1}0000000000000000$
       $O\ NC\ O\ NO\ NO\ N$
       $0\ a\ l\ q\ a\ l\ a\ mu$

## Discussion

The work described in this paper was based on the Optimality Theory of Prince and Smolensky (1993), making three additional assumptions:

1. All constraints are binary, or can be recast as binary constraints. This seems to be true of all constraints used by P&S.

2. That the initial set of candidates, the output of GEN, is a regular set which can be specified by a finite-state automaton.

3. Each constraint can be implemented as a regular transducer which determines the list of marks for each candidate.

On the basis of these assumptions, the following developments were made:

- Transducers were defined which computed not just a single constraint, but an ordered hierarchy of constraints.

- An algorithm for a product operation on these transducers was given. With this operation transducers representing constraints could be applied to sets of candidates, and also be combined into transducers representing collections of constraints.

- Algorithms were presented for
   - finding the harmony of the optimal candidate in a transducer, and
   - culling all non-optimal paths from a transducer.

- These algorithms were proved to fulfill their goals.

- The worst-case complexity of the combined algorithm in terms of harmony comparisons was found to be less than $o(|arcs|\log|states|)$, for a given transducer.

Using the assumptions and algorithms given here, there are three stages to computing a derivation in OT:

1. Specify the regular class of candidates as an automaton.

2. Build up the product of this automaton with the transducers of each constraint in decreasing order of priority.

3. Cull suboptimal paths.

There are a three more points worth noting. Firstly, the constraints in a hierarchy can be precompiled into a single transducer. Each application to a set of candidates then only requires a single product operation followed by a cull.

Secondly, casting the output of GEN and all constraints as regular means that, at all stages in a derivation, the set of candidates is regular. This is because the output of the product and culling operations are regular — both return automata.

Finally, this specification of OT in terms of regular sets and finite-state automata opens the way for more rigorous exploration of the differences between OT and declarative phonological theories, such as One-Level Phonology (Bird and Ellison 1994), which is a constraint-based phonology that defines inviolable constraints with automata.

# References

Bird, S. (1994). *Computational Phonology: A Constraint-Based Approach*. Studies in Natural Language Processing. Cambridge University Press.

Bird, S. & Ellison, T. M. (1994). One level phonology: autosegmental representations and rules as finite automata. *Computational Linguistics*, *20*, 55–90.

Ellison, T. M. (1994). Constraints, Exceptions and Representations. In *Proceedings of the First Meeting of the ACL Special Interest Group in Computational Phonology*, (pp. 25–32). ACL.

Itô, J. & Mester, R. A. (1993). Licensed segments and safe paths. *Canadian Journal of Linguistics*, *38*, 197–213.

McCarthy, J. (1993). A case of surface constraint violation. *Canadian Journal of Linguistics*, *38*, 169–195.

McCarthy, J. & Prince, A. (1993). Prosodic Morphology I — Constraint Interaction and Satisfaction. Unpublished Report.

Prince, A. S. & Smolensky, P. (1993). Optimality Theory: Constraint Interaction in Generative Grammar. Technical Report 2, Center for Cognitive Science, Rutgers University.

Scobbie, J. M. (1991). *Attribute-Value Phonology*. PhD thesis, University of Edinburgh.

Scobbie, J. M. (1993). Constraint violation and conflict from the perspective of declarative phonology. *Canadian Journal of Linguistics*, *38*, 155–167.